\documentclass[sigconf,nonacm]{acmart}

\usepackage{iftex}
\ifPDFTeX
  \usepackage[utf8]{inputenc}
  \usepackage[T1]{fontenc}
\fi
\usepackage{booktabs}
\usepackage{amsfonts}
\usepackage{nicefrac}
\usepackage{microtype}
\usepackage{xcolor}
\usepackage{graphicx}
\usepackage{amsmath}

\renewcommand\footnotetextcopyrightpermission[1]{}
\title{Gen Through the Lens: Camera-Controllable Video Generation via Cross-Coordinate Conditioning Fusion}

\author{Shuai Li\textsuperscript{*}, Jinxu Xiang\textsuperscript{*}, Xinyuan Li, Hanlei Guo, Xiyang Tan, Alan Zhao, Xiaotong Zhao, Ran Zhang\rlap{\textsuperscript{\textdagger}}
}
\affiliation{\institution{Tencent Video Solar System}\country{}}

\newcommand{\parsection}[1]{\noindent\textbf{#1.}~}

\begin{document}

\begin{abstract}
We present Gen Through the Lens, a camera-controllable image-to-video generation framework that synthesizes videos from a given reference image while enabling realistic and physically consistent camera operations. Unlike previous models that treat camera motion as abstract conditioning or rely on a single coordinate system, our approach explicitly decomposes and fuses control signals across three coordinate spaces: the camera coordinate (modeled by intrinsic parameters), the world coordinate (represented by spatial geometry hints), and the object coordinate (described by 3D bounding boxes of target subjects). Through cross-coordinate conditioning fusion, these heterogeneous cues are unified to guide the generative process, enabling precise control of camera motion, focus, and framing while preserving temporal and spatial coherence. Extensive experiments demonstrate that our method produces controllable and visually consistent results across diverse camera operations, outperforming existing approaches in both realism and editability.
\end{abstract}

\begin{teaserfigure}
    \centering
    \includegraphics[width=1\linewidth]{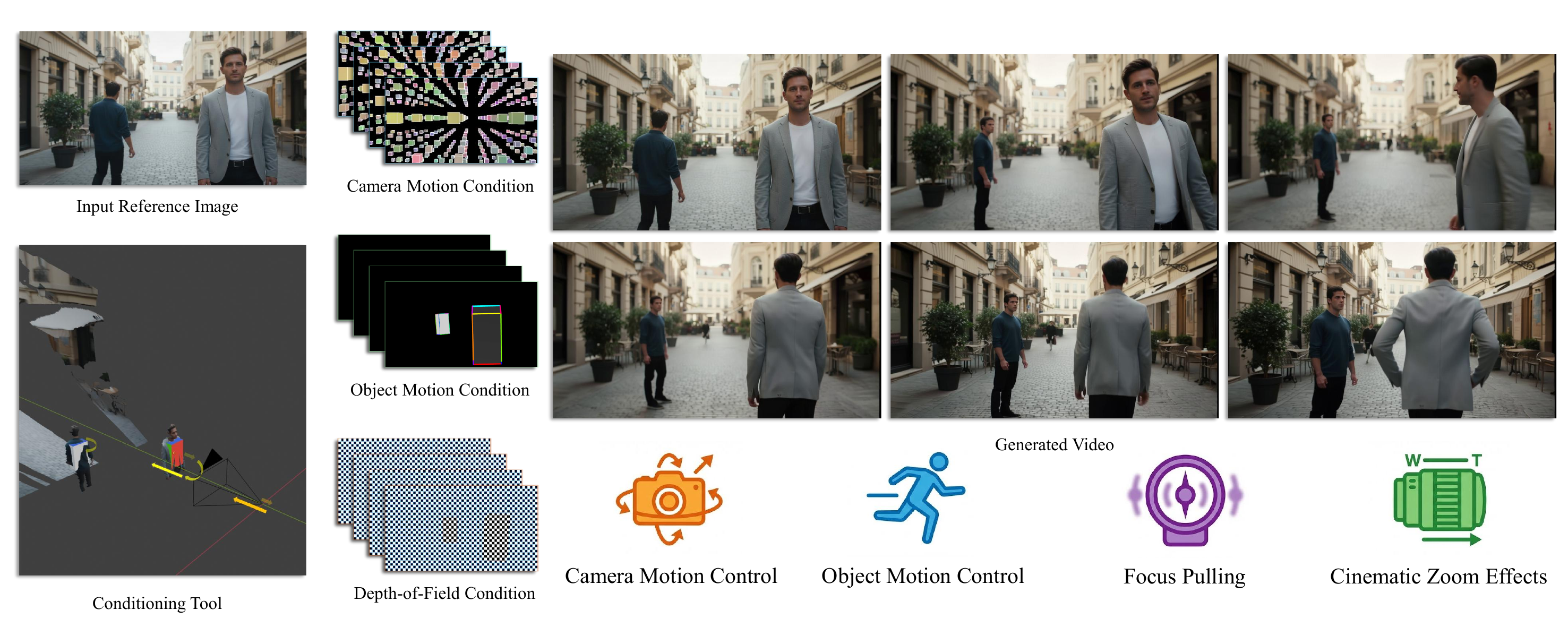}
    \caption{Overview of \textbf{Gen Through the Lens}, the first image-to-video generation framework that jointly enables controllable camera motion, object motion, and depth-of-field effects.
Starting from a single input image, our model supports cinematic camera trajectories, zoom and focus transitions, and stable object dynamics in a physically consistent manner.
The lower left shows a modeling software environment for editing camera trajectories, object motion, and depth-of-field information.
The second column shows the three complementary conditioning signals, and the right side shows the generated video results, with the model’s key controllable capabilities summarized below.
}
    \label{fig:teaser}
\end{teaserfigure}

\maketitle
\renewcommand{\shortauthors}{Li et al.}
\begingroup
\renewcommand{\thefootnote}{\fnsymbol{footnote}}
\footnotetext[1]{Equal contribution.\hspace{2em}\textsuperscript{\textdagger} Corresponding author.}
\endgroup

\section{Introduction}
\label{sec:intro}

Recent advances in video generation have achieved impressive visual realism, yet providing precise and physically meaningful control over camera behavior remains a fundamental challenge.
Most existing generative models either treat camera motion as an implicit byproduct of appearance statistics or encode it using low-level signals such as optical flow or relative pose embeddings, making it difficult to specify complex camera operations such as metric-scale motion, depth-of-field variation, or coupled changes in camera intrinsics and extrinsics.
A key difficulty lies in the heterogeneous nature of camera-related information, which simultaneously involves intrinsic parameters (e.g., focal length and distortion), extrinsic motion in 3D space, and interactions with scene geometry and object motion.
Collapsing these factors into a single coordinate system or abstract latent codes obscures their physical meaning and limits controllability, particularly for advanced cinematic effects such as depth-of-field control or the Hitchcock zoom~\cite{hitchcock1958vertigo}.

In this work, we present \textbf{Gen Through the Lens} (GTL), an image-to-video generation framework that enables explicit and physically grounded control over camera motion, object motion, and depth-of-field by decomposing and fusing camera-related signals across multiple coordinate spaces.
Our approach introduces a cross-coordinate conditioning formulation that jointly models
(i) the \emph{camera coordinate}, capturing intrinsic parameters and lens effects,
(ii) the \emph{world coordinate}, encoding metric-scale camera motion and spatial geometry, and
(iii) the \emph{object coordinate}, representing object motion through 6D bounding boxes parameterized by 3D translation and 3D rotation.
By preserving the physical meaning of each coordinate space and explicitly fusing them during generation, our framework enables precise and interpretable control over camera motion, focus, and framing.

Building upon this formulation, we design three complementary conditioning signals corresponding to the three coordinate spaces:
(1) a \emph{camera motion condition}, which visualizes metric-scale camera extrinsics and intrinsics in the world coordinate space and enables explicit control over camera translation, rotation, and zoom;
(2) an \emph{object motion condition}, which represents foreground dynamics using 3D bounding boxes with 6D pose, thereby enabling disentangled control over object translation and rotation;
and (3) a \emph{depth-of-field condition}, which encodes focus depth and defocus behavior using layered depth cues, allowing explicit and time-varying control of bokeh effects.
These conditions are geometrically consistent and mutually complementary, jointly enabling controllable video synthesis with coherent camera motion, stable object dynamics, and physically meaningful depth-of-field behavior.

However, effectively leveraging physically grounded conditions requires accurate and metrically consistent camera supervision, which is difficult to obtain from unconstrained videos due to scale ambiguity, temporal instability, and intrinsic–extrinsic inconsistencies.
To address this challenge, we introduce a metric-scale dataset labeling pipeline that combines relative camera pose estimation with monocular metric depth and robust scale recovery, constructing a unified metric camera space for reliable camera-controllable supervision.

Extensive quantitative evaluations demonstrate that our method significantly improves camera controllability, geometric consistency, and motion stability compared to state-of-the-art approaches.
In addition, qualitative results showcase compelling cinematic applications, including controllable depth-of-field effects and stable synthesis of the Hitchcock zoom.
Together, these results highlight the importance of explicit, cross-coordinate camera modeling for controllable and physically consistent video generation.

\section{Related Work}
\label{sec:related}

\subsection{Camera-Controllable Video Generation}
Camera motion control in generative video has evolved from coarse motion cues to explicit trajectory conditioning. Early text-based approaches such as AnimateDiff~\cite{guo2023animatediff} and MotionDirector~\cite{zhao2024motiondirector} typically fine-tune LoRA modules for a small set of motion patterns, providing convenient but imprecise control. Recent trajectory-guided methods instead condition video diffusion on camera parameters. MotionCtrl \cite{wang2024motionctrl} conditions generation on raw camera parameters repeated over the spatial grid, while CameraCtrl~\cite{he2024cameractrl} and CamCo~\cite{xu2024camco} encode camera parameters as pixel-wise Pl\"ucker ray embeddings to obtain more discriminative viewpoint control. Several works lift the reference image to 3D and render proxy views for conditioning, such as ViewCrafter~\cite{yu2024viewcrafter} and I2VControl-Camera~\cite{feng2024i2vcontrol}, improving spatial coherence under viewpoint changes. To better handle long-range motion and cross-frame geometry, CamI2V~\cite{zheng2024cami2v} introduces epipolar attention mechanisms, and RealCam-I2V~\cite{li2025realcam} further targets real-world usability by aligning trajectories to metric scale. Beyond pure conditioning, GEN3C~\cite{ren2025gen3c} strengthens controllability by maintaining a 3D cache rendered along user-specified trajectories, which also improves temporal 3D consistency. Despite these advances, most trajectory-based controllers still rely on approximate geometry or relative-scale supervision and treat the camera as a simplified projection model, which can lead to scale drift, limited robustness under large rotations or occlusions, and weak support for optical phenomena beyond pose changes.

\subsection{Object-Controllable Video Generation}
Object control in generative video has progressed from coarse region constraints to explicit trajectory-driven motion specification. Early approaches primarily rely on spatial conditions to anchor object placement but provide limited control over object dynamics. Recent works instead adopt drag-style interaction, where sparse point trajectories convey user intent and are injected into diffusion models through explicit motion representations. DragNUWA~\cite{yin2023dragnuwa} introduces joint text-image-trajectory conditioning for open-domain object manipulation, while ImageConductor~\cite{li2025image} decouples camera and object motions via separate LoRA branches for more precise interactive control. 
Many methods use optical flow as an intermediate representation: sparse trajectories are converted to sparse or dense flow to guide temporal attention or feature warping, as in Motion-I2V~\cite{shi2024motion} and MOFA-Video~\cite{niu2024mofa}. 
Beyond flow, Tora~\cite{zhang2025tora} encodes trajectories into a trajectory map and integrates them into a scalable diffusion-transformer pipeline via a trajectory extractor and fusion module, and DragAnything~\cite{wu2024draganything} uses an entity representation with mask-based feature reorganization to improve object-level coherence under dragging. 
Despite improved controllability, these methods remain sensitive to trajectory placement and depth or occlusion ambiguities, can introduce local distortions under large shape variations, and often suffer from control conflicts when object motion and global camera motion are specified simultaneously.

\subsection{Collaborative Motion Control}
Collaborative motion control targets video generation with simultaneous camera motion and local object dynamics, where naive 2D motion controls often lead to local conflicts under complex camera movements.
Early approaches unify camera and object motion in a single 2D representation.
AnimateAnything~\cite{lei2024animateanything} converts both motions into a unified optical-flow field, but entangles global camera-induced displacement with object-centric motion, limiting geometric interpretability.
Subsequent works adopt point-based motion control derived from 2D perception signals. 
MotionPrompting~\cite{geng2024motionprompting} and I2VControl-Camera~\cite{feng2024i2vcontrol} leverage dense point embeddings to guide motion generation, but collaborative control remains rooted in 2D cues and is typically specified at region or segmentation-mask granularity, restricting fine-grained camera–object interaction modeling.
Perception-as-Control~\cite{chen2025perception} introduces a 3D-aware motion representation with separate encoders for camera and object control to reduce local conflicts. 
Despite improved motion harmony, collaborative approaches still depend on upstream perception quality, simplified 3D surrogates may miss motion factors, and overall controllability remains constrained by the underlying generator.

\section{Method}
\label{sec:method}

In this section, we present a metric-scale supervision and conditioning pipeline that first recovers semantic, camera, object, and depth-related supervision from real-world videos, and then constructs geometry-aware conditioning signals for controllable video generation.
Together, these components enable disentangled yet coherent control over camera motion, object motion, depth-of-field, and zoom. The overall data processing workflow is illustrated in Fig.~\ref{fig:data_labeling}.
Detailed implementation of each component is provided in Appendix Sec.~\ref{sec:appendix_conditions}. 

\begin{figure}[t]
\centering
\includegraphics[width=\linewidth]{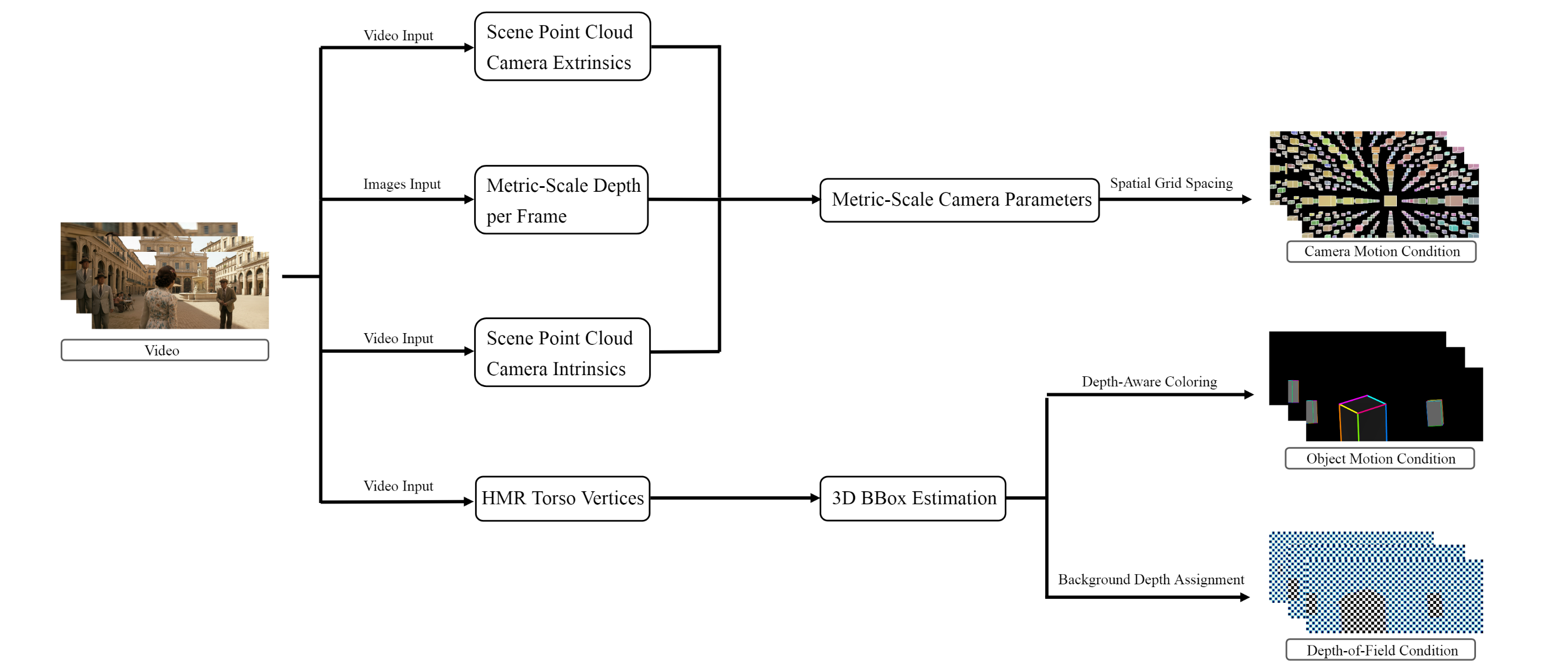}
\caption{Overview of the data processing pipeline. Recovered metric-scale camera parameters and human-centered 3D geometry are converted into geometry-aware conditions for camera motion, object motion, and depth-of-field control.}
\label{fig:data_labeling}
\end{figure}

\subsection{Label Prompt}

We use InternVL2.5~\cite{chen2024expanding} to generate dense textual descriptions for each video, which serve as semantic supervision during training.
To avoid information leakage and ensure disentanglement between camera control and content description, we explicitly constrain the prompt to exclude camera-related information, such as camera motion, viewpoint, and shooting angles.
The prompt is designed to encourage temporally coherent, detailed, and factually accurate descriptions that focus on scene content and object dynamics.

\subsection{Camera Modeling}

Metric-scale camera extrinsics and accurate camera intrinsics are essential for injecting reliable \emph{camera motion condition} into camera-controllable generative models.
Wan2.1~\cite{wan2025wan} is trained on a RealEstate-style dataset~\cite{zhou2018stereo} that effectively normalizes camera trajectories under a fixed foreground depth assumption; when this assumption is violated at inference, the generated motion exhibits noticeable scale inconsistencies.
To address this issue, we build a labeling pipeline that recovers metric-scale camera extrinsics and intrinsics for each video, forming a unified \emph{metric camera space}.

Accurate camera-controllable video generation requires camera extrinsics and intrinsics that are both geometrically reliable and consistent under metric scale.
However, existing methods typically specialize in only part of this problem, such as temporally stable camera motion, per-frame metric depth, or accurate intrinsic estimation.
We therefore combine three complementary approaches—Pi3~\cite{wang2025pi}, Moge2~\cite{wang2025moge}, and MapAnything~\cite{keetha2025mapanything}—to recover metric-consistent camera parameters.
Specifically, Pi3 provides temporally stable camera extrinsics up to an unknown global scale, Moge2 predicts per-frame metric depth, and MapAnything provides reliable camera intrinsics.

We first estimate a robust global scale factor from Pi3 and Moge2 using pairwise 3D distance ratios across the video, which resolves the scale ambiguity in camera extrinsics.
We then align the camera intrinsics predicted by MapAnything to the same metric space using a consistent scaling strategy, ensuring geometrically accurate projection across frames.
The resulting metric-scale camera parameters are used as supervision for training.
Quantitative evaluation of the recovered camera parameters is reported in Table~\ref{tab:camera_accuracy}.

Based on the recovered parameters, we construct a \emph{camera motion condition} using an intermediate world-coordinate representation rather than raw Pl\"ucker embeddings~\cite{plucker1865xvii} or per-frame point clouds.
We normalize camera poses to the first frame and represent camera trajectories using a structured grid in world coordinates, which is rendered with the recovered intrinsics to ensure geometric consistency across frames.
This representation encodes camera motion in a structured and controllable form.

\begin{figure*}[t!]
\centering
\includegraphics[width=0.9\textwidth]{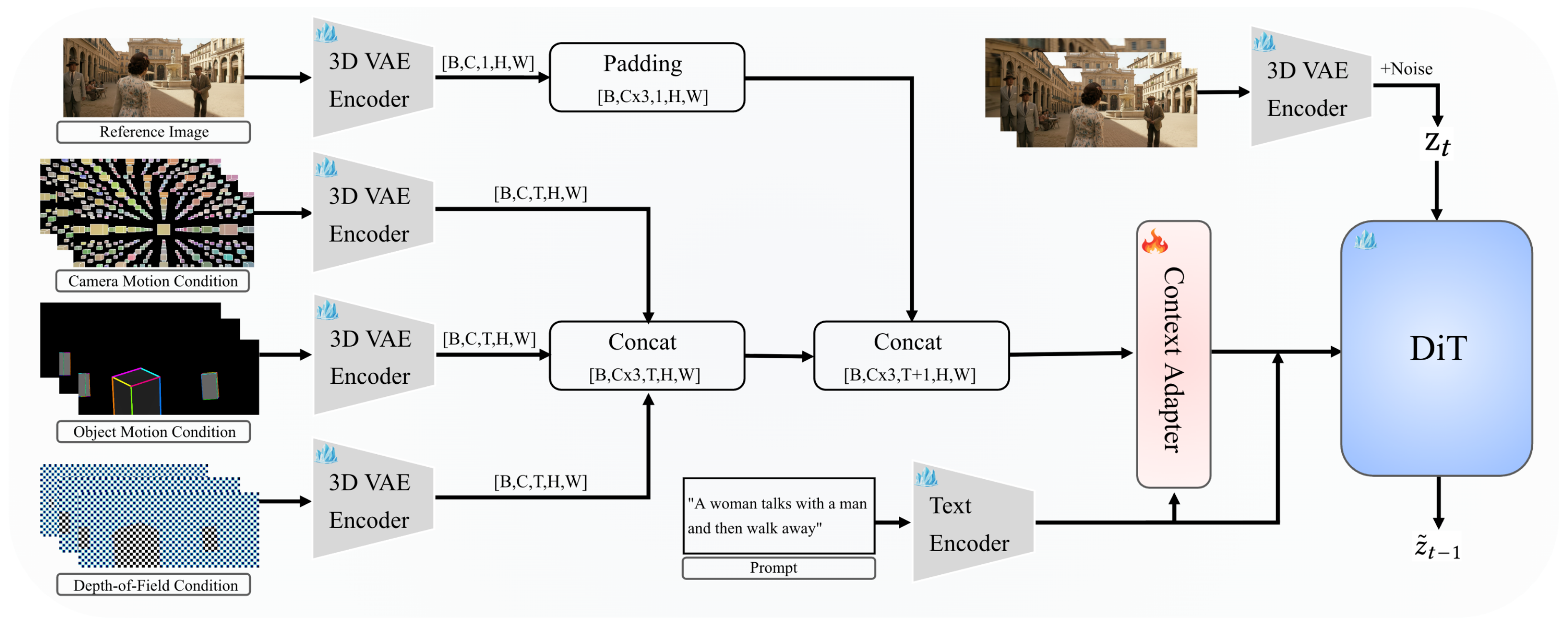}
\caption{
Overview of the proposed framework.
The model jointly conditions video generation on camera motion, object motion, and depth-of-field signals across camera, world, and object coordinate spaces.
}
\label{fig:overview}
\end{figure*}

\subsection{3D Bounding Box Modeling}

Controlling object motion solely from 2D image coordinates inevitably discards spatial information, including depth ordering under occlusion and object orientation.
This ambiguity motivates the need for an explicit \emph{object motion condition}.

Following Cinemaster~\cite{wang2025cinemaster}, we represent object motion using 3D bounding boxes in full 6D, including 3D translation and 3D rotation.
We focus on human subjects, as their geometry and motion can be reliably reconstructed from monocular videos using off-the-shelf Human Mesh Recovery (HMR) methods, enabling scalable automatic annotation.
We employ TokenHMR~\cite{dwivedi2024tokenhmr} for identity-consistent tracking, and fit bounding boxes only to the rigid torso vertices of the recovered SMPL mesh~\cite{loper2023smpl}.
This design mitigates large shape variations caused by articulated limbs or head motion (e.g., bending or sitting), resulting in more stable bounding boxes and temporally consistent 6D object motion supervision.

We then convert the recovered 3D boxes into an \emph{object motion condition} that provides explicit local spatial cues.
Complementing the camera motion condition, which defines the global metric space, the 3D bounding box condition indicates precise screen-space occupancy, relative depth ordering, and orientation.
The condition is rendered as a projected 3D bounding box with depth-aware filling and consistent edge coloring, enabling the model to infer object rotation and placement without explicit angle representations.
As shown in Fig.~\ref{fig:bbox_figure}, although the model is trained only on humans, the learned object motion condition generalizes well to generic non-human objects.

\begin{table}[t]
\centering
\setlength{\tabcolsep}{6pt}
{\scriptsize
\begin{tabular}{lccc}
\toprule
Method
& \begin{tabular}[c]{@{}c@{}}Cam Trans.\\ Err. $\downarrow$\end{tabular}
& \begin{tabular}[c]{@{}c@{}}Cam Rot.\\ Err. (°) $\downarrow$\end{tabular} \\
\midrule
VGGT         & 41.25 & 20.34 \\
Pi3           & 40.36 & 17.21 \\
MapAnything  & 12.39 & 17.43 \\
\textbf{Ours}        & \textbf{11.74} & \textbf{17.11} \\
\bottomrule
\end{tabular}
}
\caption{Metric-scale camera pose accuracy on the MultiCamVideo Dataset with ground-truth camera annotations.
Methods that estimate camera poses only up to scale (VGGT and Pi3) exhibit large errors on sequences with significant camera motion.
Our method recovers metrically consistent camera parameters and achieves lower errors, validating the effectiveness of the proposed pipeline.}
\label{tab:camera_accuracy}
\end{table}

\subsection{Depth-of-Field Modeling}

To model depth-of-field effects, we construct \emph{depth-of-field condition} signals using recovered metric depth and a physically motivated bokeh operator following AkiRa~\cite{wang2025akira}, where blur is determined by the deviation between pixel depth and the focal plane.
We use the same operator for both bokeh-augmented data generation and condition construction, ensuring consistency between training signals and supervision.
The recovered metric camera space enables flexible focus modeling.
By specifying time-varying focus depths, we can synthesize diverse behaviors such as smooth focus transitions and subject tracking.
To improve temporal stability, we apply a boundary-aware smoothing step to mitigate artifacts near large depth discontinuities.

For conditioning, we adopt a layered representation instead of directly encoding pixel-wise focus distance.
We approximate unknown background depth using a small set of canonical depth planes and combine them with foreground depth derived from 3D bounding boxes.
The resulting representation captures the interaction between focus distance and scene depth while avoiding reliance on accurate full-scene depth estimation.

\subsection{Zoom Video Augmentation}
In practical cinematography, zoom is an essential component of camera behavior, yet recovering time-varying focal lengths directly from unconstrained videos is highly unreliable, and such sequences are difficult to curate at scale.
We therefore synthesize zoom effects through data augmentation.
Specifically, we model focal-length changes by applying a temporal center crop with randomly sampled crop factors and zoom intervals, and resizing the result back to the native resolution to simulate push-in and pull-out shots while keeping the final image resolution fixed.
Once the zoom parameters are sampled for a video, the same transformation is consistently applied to all conditioning streams—including the camera motion condition video, the object motion condition video, and the depth-of-field condition video—to preserve pixel-wise alignment.

\section{Experiments}
\label{sec:exp}

\raggedbottom
\begin{table*}[t]
\centering
\setlength{\tabcolsep}{4.5pt}
{\scriptsize
\begin{tabular}{lccccccccccc}
\toprule
Method
& FID $\downarrow$
& FVD $\downarrow$
& \begin{tabular}[c]{@{}c@{}}Qwen VQA\\ Score $\uparrow$\end{tabular} 
& \begin{tabular}[c]{@{}c@{}}Cam Trans.\\ Err. $\downarrow$\end{tabular}
& \begin{tabular}[c]{@{}c@{}}Cam Rot.\\ Err. (°) $\downarrow$\end{tabular}
& mIoU $\uparrow$
& \begin{tabular}[c]{@{}c@{}}BBox Center\\ Err. (\%) $\downarrow$\end{tabular}
& \begin{tabular}[c]{@{}c@{}}BBox Rot.\\ Err. (°) $\downarrow$\end{tabular}
& \begin{tabular}[c]{@{}c@{}}Traj.\\ Err. (\%) $\downarrow$\end{tabular} \\
\midrule
PAC         &  70.17 &  172.51 &  0.56 &  18.15  &  8.77  & 0.52  & 6.37  & 41.83  &  8.68 \\
RC-I2V & 55.12  &  98.76 & 0.70  &  9.86  &  \textbf{4.63}  & $0.45^{\ast}$  &  $7.42^\ast$ & $47.34^\ast$  &  7.29 \\
GEN3C       & 64.24  & 114.33  &  0.62 &  11.38  &  5.02  & $0.47^{\ast}$  & $8.24^\ast$  &  $46.22^\ast$ &  8.39 \\
\midrule
\textbf{GTL 14B}      & \textbf{53.71} & \textbf{86.54} & \textbf{0.81} & \textbf{8.59} & 4.89 & \textbf{0.61} & \textbf{4.40} & \textbf{21.73} & \textbf{6.37} \\
\bottomrule
\end{tabular}
}
\caption{
Quantitative comparison with state-of-the-art methods.
Lower is better ($\downarrow$) for error-based metrics and latent space distance, while higher is better ($\uparrow$) for overlap and semantic metrics.
Screen-space errors are reported as percentages of the image diagonal.
Metrics marked with $\ast$ should be interpreted with caution, as RealCam-I2V and GEN3C do not support explicit control of object motion or target object positioning, which may bias object-related metrics toward worse values.
}
\label{tab:quantitative}
\end{table*}

\subsection{Training Paradigm}
\label{sec:training}

We curate a dataset with significant camera and object dynamics from DynPose-100K~\cite{rockwell2025dynamic}, MultiCamVideo~\cite{bai2025recammaster}, BEDLAM2.0~\cite{tesch2025bedlam2}, and internal data, totaling 110k samples with camera trajectories and 3D bounding box annotations. 
We further construct a zoom dataset (8k) and a bokeh dataset (15k).
All training videos are unified to 81 frames at 480p resolution.

We build GTL 14B on Wan2.1-14B~\cite{wan2025wan}, freezing the encoder, decoder, and DiT, and training only the Context Adapter.
Training follows a three-stage curriculum from camera to object and finally to bokeh control.
We initialize from partial VACE DiT weights~\cite{jiang2025vaceallinonevideocreation}.
Stage 1 trains camera control using trajectory annotations; Stage 2 adds 3D bounding boxes and zoom data for object motion and zoom; Stage 3 introduces bokeh data for joint control.
The three stages run for 11k, 13k, and 7k iterations with batch size 8 and learning rate $5 \times 10^{-5}$, totaling approximately 288 GPU hours on 8 NVIDIA H800 GPUs.
An overview of the model and training pipeline is shown in Fig.~\ref{fig:overview}.

\subsection{Evaluation Metrics}
\label{sec:metrics}
We present a comprehensive evaluation of our method across multiple complementary aspects, including camera controllability, geometric consistency, motion fidelity, and visual quality.
For metrics that rely on extracted signals, we apply the same extraction pipeline to both the real input videos and the generated videos, and compare the recovered features, using explicit ground truth when available and real-video signals as pseudo-ground-truth otherwise.
Brief metric descriptions are given below, with full details provided in Appendix~\ref{sec:metrics_appendix}.

\parsection{Camera motion accuracy}
Camera controllability is evaluated using two metrics:
(i) camera translation error, which quantifies relative deviation of the camera trajectory in 3D space;
and (ii) camera rotation error, which measures orientation accuracy using geodesic distance on $\mathrm{SO}(3)$.
All camera trajectories are normalized to the first frame to ensure a consistent reference across sequences.

\parsection{Object-level geometric stability}
To assess foreground consistency under camera motion, we report the mean Intersection-over-Union (mIoU) measured by SAM2~\cite{ravi2024sam} between predicted and reference foreground masks.
In addition, we evaluate 3D bounding box pose accuracy using both screen-space center deviation and 3D rotational error, which measure how accurately the generated videos follow the object controls.

\parsection{Trajectory-based motion accuracy}
We assess motion fidelity using a trajectory consistency error computed from point tracks extracted by CoTracker~\cite{karaev2024cotracker}.
This metric measures the alignment between induced 2D motion and the reference video by comparing point trajectories across frames, and reflects the combined effects of camera and object motion.
When evaluated over longer trajectories, it becomes sensitive to accumulated drift, serving as an indicator of long-term motion stability.

\parsection{Visual quality and semantic alignment}
We report Fr\'echet Inception Distance (FID)~\cite{heusel2017gans} and Fr\'echet Video Distance (FVD)~\cite{unterthiner2019fvd} to evaluate perceptual realism and temporal coherence.
To assess semantic alignment with text prompts, we additionally report the Qwen VQA Score~\cite{wang2024qwen2}, which measures the consistency between generated videos and input prompts using a vision--language model.

\subsection{Evaluation Datasets}
\label{sec:eval_datasets}

We evaluate our method on a held-out test set consisting of 1k video clips drawn from DynPose-100K, the MultiCamVideo Dataset, BEDLAM2.0, and our proprietary internal data.
Together, these datasets cover a broad spectrum of foreground occupancy ratios, camera motion complexity, and data sources, including real-world daily-life recordings, cinematic videos, and synthesized scene videos.
All evaluation videos are disjoint from the training set.
Videos are processed using the same metric-scale annotation pipeline described in Sec.~\ref{sec:method}, producing camera parameters, foreground masks, and 3D bounding box poses.
This unified annotation enables consistent and fair comparison across different methods under identical evaluation conditions.
Unless otherwise specified, all quantitative results reported later are computed on this evaluation set.

\subsection{Camera Pose Estimation Accuracy}
\label{sec:camera_accuracy}

\begin{figure*}[t!]
\centering
\includegraphics[width=0.95\textwidth]{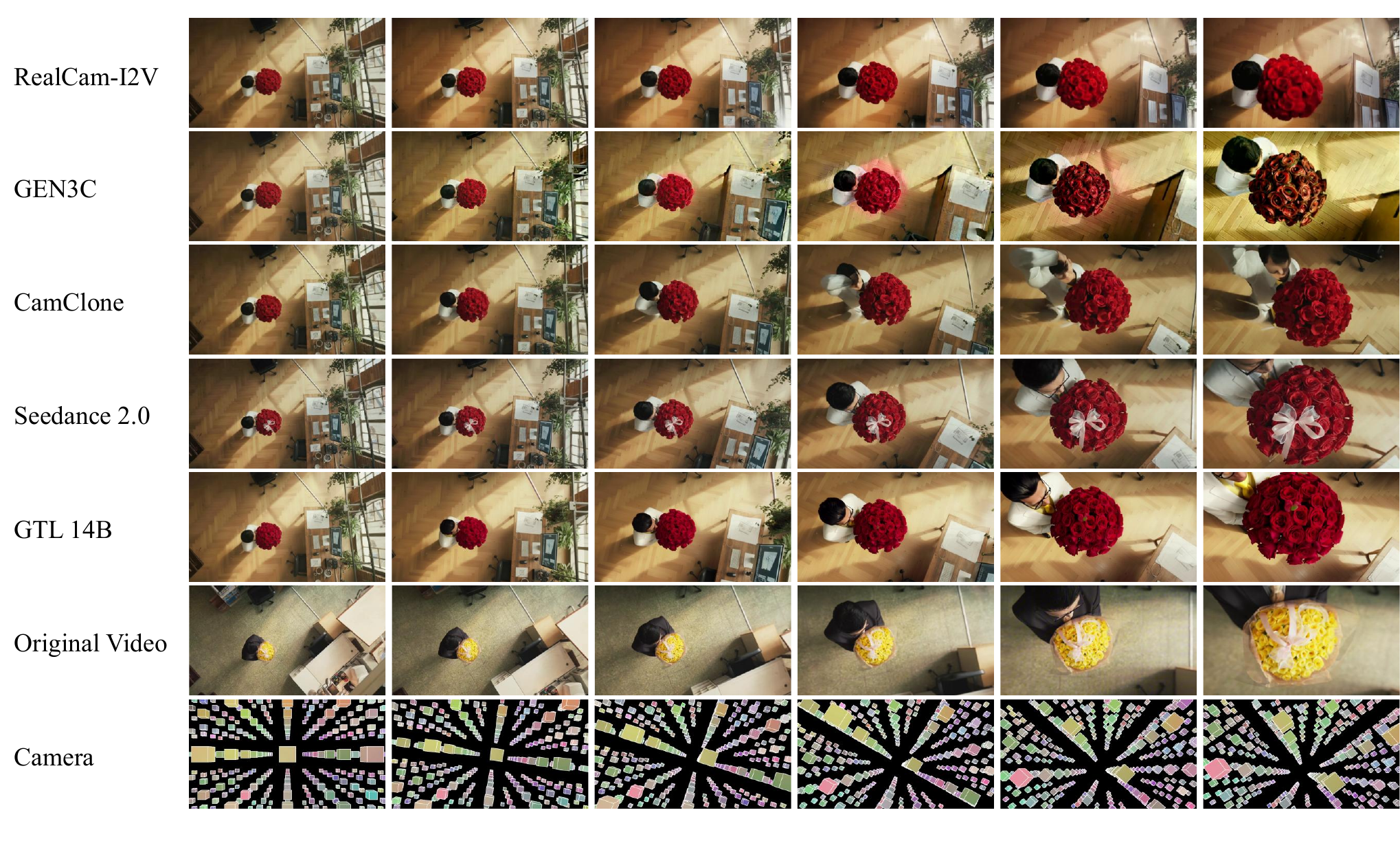}
\caption{
 Qualitative comparison of camera controllability with state-of-the-art methods. Gen3C and RealCam-I2V use identical camera motion extracted from the original video, with the first frame as input. CamCloneMaster and Seedance 2.0 use the original video as the camera reference. Our Camera condition encodes recovered extrinsics and intrinsics.
 For clarity, text prompts are omitted in this figure and provided in the Appendix Sec.~\ref{sec:camera_prompt}.
}
\label{fig:main_figure}
\end{figure*}

Subsequent camera-related evaluations rely on metric-scale camera poses extracted from generated videos.
To ensure that reported errors reflect generation quality rather than estimation inaccuracies, we first assess the accuracy of our camera pose extraction pipeline. 
We compare our method with representative approaches, including VGGT~\cite{wang2025vggt}, Pi3~\cite{wang2025pi}, and MapAnything~\cite{keetha2025mapanything}, using camera translation and rotation errors. 
As shown in Table~\ref{tab:camera_accuracy}, our method achieves improved metric-scale accuracy, validating the effectiveness of the proposed pipeline for reliable supervision. 
We further provide qualitative validation by visualizing the extracted camera motion condition alongside the input videos.
As illustrated in Fig.~\ref{fig:main_figure}, the recovered motion faithfully reflects scene structure and camera dynamics, with consistent changes in position and orientation, making it directly interpretable.
An ablation study comparing training with camera parameters from MapAnything and our extraction pipeline is provided in Appendix Table~\ref{tab:ablation_dataset}.

\begin{figure*}[t!]
\centering
\includegraphics[width=0.95\textwidth]{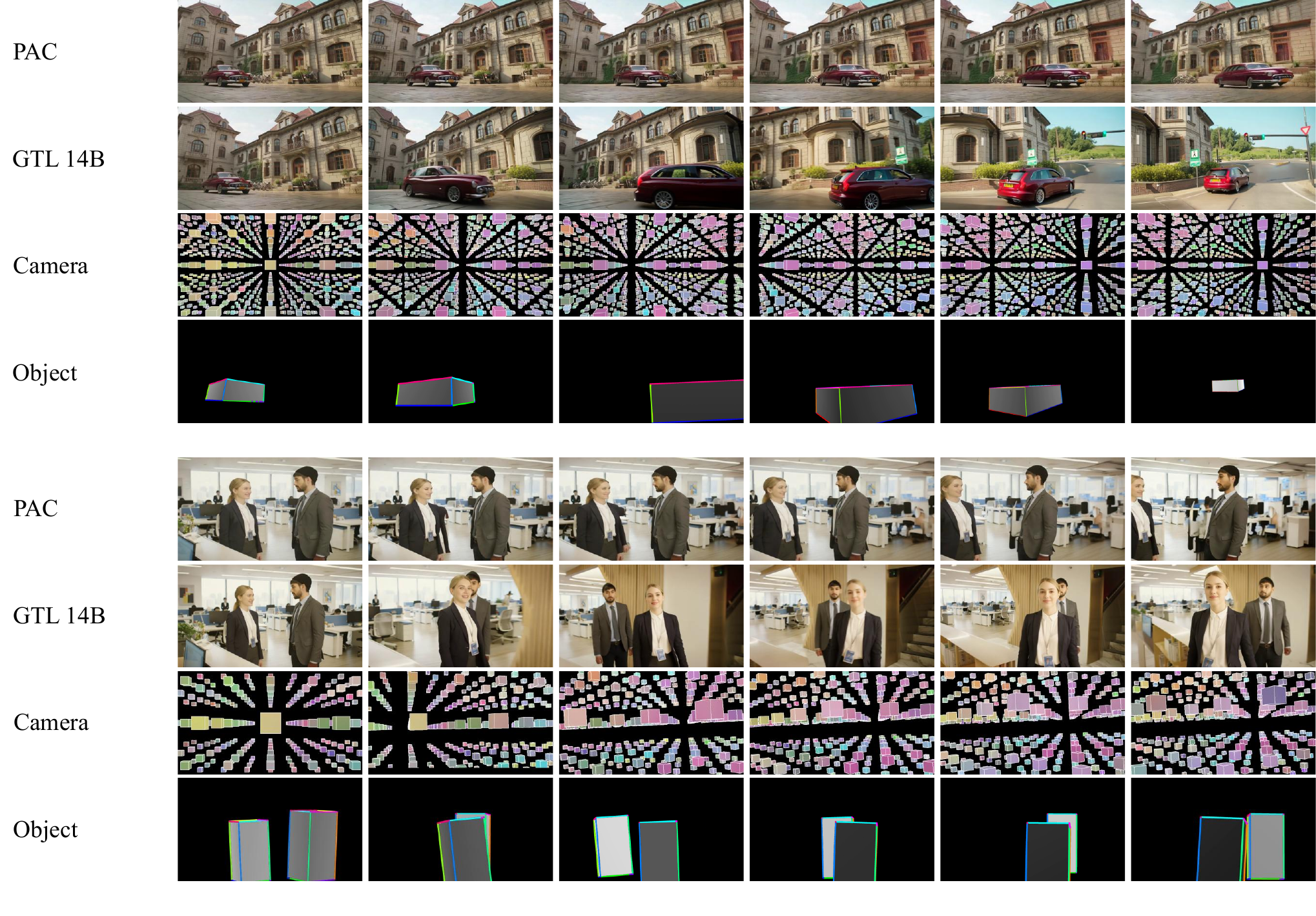}
\caption{
 Qualitative comparison of object motion control. We compare our method with Perception-as-Control under aligned camera conditions and comparable object motion controls, focusing on the controllability and stability of foreground object motion.
 For clarity, text prompts are omitted in this figure and provided in the Appendix Sec.~\ref{sec:bbox_prompt}.
}
\label{fig:bbox_figure}
\end{figure*}

\begin{figure*}[t!]
\centering
\includegraphics[width=1.0\textwidth]{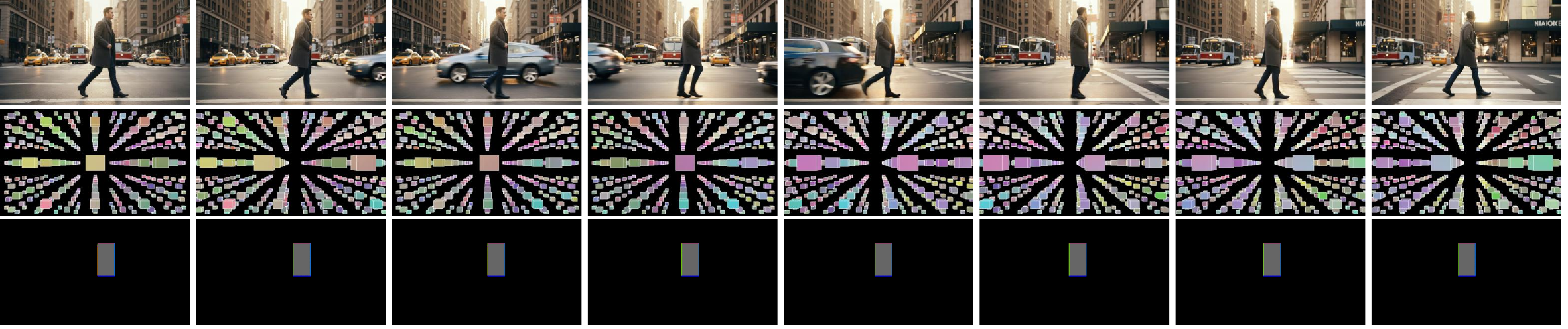}
\caption{Disentanglement between camera and object motion. 
Keeping the object fixed in screen space under camera motion requires the object to move together with the camera in 3D. 
The prompts used here avoid specifying object motion and are provided in Appendix Sec.~\ref{sec:disentangle_prompt}.}
\label{fig:disentangle}
\end{figure*}

\subsection{Comparison with State-of-the-Art}
\label{sec:quantitative}

Having established the accuracy of our camera pose extraction pipeline, we evaluate the proposed framework GTL 14B against representative state-of-the-art methods, including Perception-as-Control (PAC)~\cite{chen2025perception}, RealCam-I2V~\cite{li2025realcam}, and GEN3C~\cite{ren2025gen3c}.
Quantitative comparisons across camera motion accuracy, object-level stability, trajectory consistency, and visual quality are reported in Table~\ref{tab:quantitative}.
The inference hyperparameters for GTL 14B are provided in Appendix Sec.~\ref{sec:hyperparameters}.

As summarized in Table~\ref{tab:quantitative}, GTL 14B achieves lower camera translation and rotation errors, indicating more accurate camera motion recovery under metric scale.
With explicit 3D bounding box conditioning, the model better regulates object translation and rotation, reducing object-centric errors.
Joint camera-object control further improves motion stability and trajectory consistency, suggesting that explicit geometry-aware conditions provide stronger motion supervision than implicit image-space guidance alone.
Perceptual and semantic metrics further show that these geometric improvements do not come at the cost of visual quality.

Recent works such as CamCloneMaster~\cite{luo2025camclonemaster} require a reference video to specify camera motion.
In our setting, this would require using the ground-truth video as input, which introduces an unfair advantage.
We therefore exclude it from quantitative evaluation and report qualitative results in Fig.~\ref{fig:main_figure}.
We also include results from Seedance 2.0~\cite{seedance2026seedance}.
Due to its less explicit separation between reference and target generation, it often transfers appearance details from the reference video (e.g., the knot on the flower), leading to inconsistencies with the intended control signal.

The qualitative results in the following figures corroborate these findings.
Fig.~\ref{fig:main_figure} shows that, when using only camera motion conditions, our method produces stable and geometrically consistent camera behavior under challenging camera trajectories.
Fig.~\ref{fig:bbox_figure} further demonstrates that jointly applying camera and object motion conditions yields consistent camera motion and well-controlled object dynamics.
Despite being trained primarily on human-centered motion supervision, the learned object motion condition generalizes effectively to non-human objects and complex scenes.
Additional qualitative results are shown in Fig.~\ref{fig:add_figure}.

To further validate disentanglement between camera and object motion, we visualize a controlled scenario in Fig.~\ref{fig:disentangle}.
Keeping the object fixed in screen space under camera motion requires the object to move together with the camera in 3D.
This demonstrates that our model disentangles apparent 2D motion from physical 3D object motion, enabling independent control of camera behavior and object dynamics.

\subsection{Ablation Study}
\label{sec:ablation}

To isolate the effect of individual design choices, we conduct ablation studies using a smaller Wan2.1-1.3B model trained on the same dataset for 30K steps under identical settings.
Although the 1.3B model exhibits slightly lower visual fidelity than the 14B model, it provides a reliable and efficient proxy for comparing relative performance across different conditioning designs.
The ablation studies focus on relative metric differences rather than absolute generation quality, which are largely unaffected by the backbone scale.
Additional ablation studies are provided in Appendix Sec.~\ref{sec:add_ablation}.

\parsection{Effect of Camera Condition Representation}
We study the impact of different camera condition representations on camera controllability.
As an alternative representation, we consider a Pl\"ucker ray map \cite{plucker1865xvii}, where camera motion is encoded using dense per-pixel rays parameterized by direction and moment.
All variants use identical camera extrinsics and intrinsics as input, differing only in the representation of the camera condition.
We evaluate controllability using camera translation and rotation errors, as reported in Table~\ref{tab:camera_ablation}.
The results show that our geometry-aware cubic grid representation provides more stable and accurate camera control than ray-based representations.

\parsection{Effect of 3D Bounding Box Conditioning}
We further analyze the role of explicit 3D bounding box conditioning in object motion control, as shown in Table~\ref{tab:object_ablation}.
Besides our proposed rotation-aware 3D bounding box representation, we consider a simplified baseline using only binary foreground masks as object-level conditions.
The mask-based representation provides only 2D spatial extent and lacks explicit depth ordering and rotational cues.
Both variants are trained with identical camera motion and depth-of-field conditions.
We evaluate object controllability using bounding box center error, bounding box rotation error, and trajectory consistency measured by CoTracker.
Results demonstrate that explicit 3D bounding box conditioning significantly improves rotational stability and long-term motion consistency.

\begin{table}[t]
\centering

\begin{minipage}{\linewidth}
\centering
\setlength{\tabcolsep}{4pt}
{\scriptsize
\begin{tabular}{lcc}
\toprule
Camera Condition
& \begin{tabular}[c]{@{}c@{}}Cam Trans.\\ Err. $\downarrow$\end{tabular}
& \begin{tabular}[c]{@{}c@{}}Cam Rot.\\ Err. (°) $\downarrow$\end{tabular} \\
\midrule
Ray Map (No Grid)  & 19.26 & 8.98 \\
\textbf{Color Cubic Grid (GTL 1.3B)}  & \textbf{10.15} & \textbf{5.07} \\
\bottomrule
\end{tabular}
}
\captionof{table}{
Ablation studies on geometry-aware conditioning.
Comparison of camera condition representations.
}
\label{tab:camera_ablation}
\end{minipage}

\vspace{0.5em}

\begin{minipage}{\linewidth}
\centering
\setlength{\tabcolsep}{4pt}
{\scriptsize
\begin{tabular}{lccc}
\toprule
Object Condition
& \begin{tabular}[c]{@{}c@{}}BBox Center\\ Err. (\%) $\downarrow$\end{tabular}
& \begin{tabular}[c]{@{}c@{}}BBox Rot.\\ Err. (°) $\downarrow$\end{tabular}
& \begin{tabular}[c]{@{}c@{}}Traj.\\ Err. (\%) $\downarrow$\end{tabular} \\
\midrule
2D Mask      & \textbf{5.94} & 45.15 & 8.37 \\
\textbf{3D BBox (GTL 1.3B)} & 6.13 & \textbf{27.96} & \textbf{7.03} \\
\bottomrule
\end{tabular}
}
\captionof{table}{
Ablation studies on geometry-aware conditioning.
Comparison of object motion conditioning.
}
\label{tab:object_ablation}
\end{minipage}

\end{table}

\begin{figure*}[t]
\centering
\includegraphics[width=0.95\linewidth]{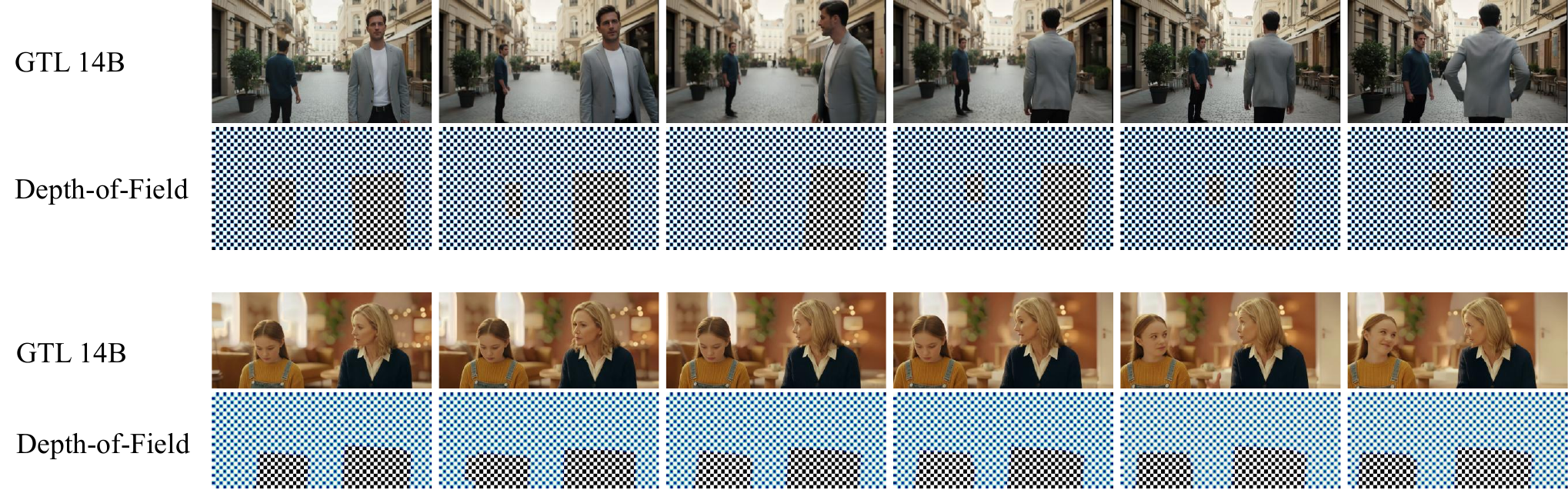}
\caption{
 Visualization of depth-of-field conditions and generated results. Our representation enables controllable bokeh strength, time-varying focus shifts, and stable depth-of-field under camera and object motion. Note that focus cues are encoded through subtle pixel-level variations, which may not be visually salient.
 The corresponding text prompts are provided in Appendix Sec.~\ref{sec:bokeh_prompt}.
}
\label{fig:bokeh_figure}
\end{figure*}

\section{Applications}
\label{sec:applications}

Our geometry-aware conditioning enables expressive cinematic effects by explicitly disentangling camera extrinsics, intrinsics, and depth-of-field.
We present two representative applications—controllable bokeh and the Hitchcock zoom—to illustrate its practical advantages.

\subsection{Controllable Depth-of-Field and Bokeh}
Our framework enables explicit and physically meaningful control over depth-of-field effects in generated videos.
By leveraging metric depth information recovered in the metric camera space, we can synthesize a wide range of bokeh behaviors with a time-varying focus depth.
In particular, the focal plane can be dynamically adjusted over time, allowing smooth focus transitions as well as focus-follow effects that track a moving subject.

Figure~\ref{fig:bokeh_figure} shows that bokeh in our model is driven by the explicit depth-of-field condition, where both the relative depth to the focal plane and the bokeh strength are integrated into the conditioning video to drive the resulting blur.
Because the focus depth is specified as a continuous, time-varying signal, our method naturally supports zoom-dependent and temporally varying focus behaviors commonly observed in real cinematography, as illustrated by the second example.

\subsection{Hitchcock Zoom}
The Hitchcock zoom \cite{hitchcock1958vertigo} is a classic cinematic effect in which the camera translates in space while adjusting focal length in the opposite direction, preserving the apparent size of the foreground subject while dramatically altering background perspective.
This coupling of camera translation and zoom poses a significant challenge for generative models, as it requires accurate disentanglement and coordinated control of camera extrinsics and intrinsics.

\begin{figure}[t]
\centering
\includegraphics[width=1.00\linewidth]{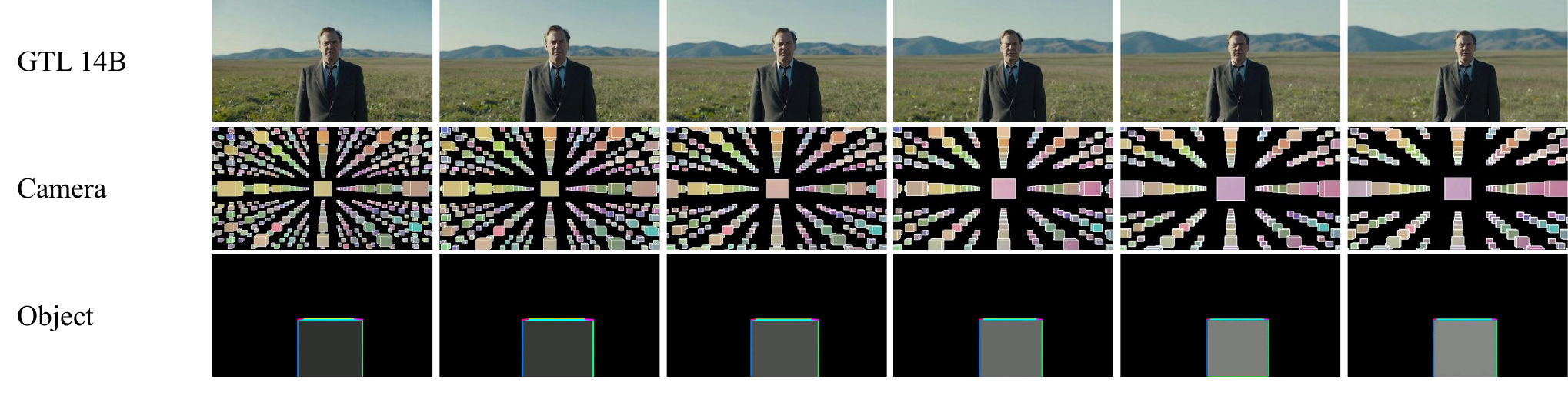}
\caption{
 Hitchcock zoom effect synthesized by our model. In this example, the camera moves backward by 3 meters while the zoom is adjusted accordingly to approximately preserve the projected size of the subject in the image, as indicated by the object motion condition video.
 The corresponding text prompts are provided in Appendix Sec.~\ref{sec:hitchcock_prompt}.
}
\label{fig:hitchcock_figure}
\end{figure}

As illustrated in Fig.~\ref{fig:hitchcock_figure}, our framework explicitly encodes camera translation and focal length variation in the camera condition.
Consequently, even when forward camera motion and inverse zoom produce visually similar image content, they correspond to distinct patterns in our camera condition, enabling stable and faithful synthesis of the Hitchcock zoom.

\section{Conclusion}
In this paper, we present Gen Through the Lens, a camera-controllable image-to-video generation framework that enables explicit and physically grounded control over camera motion, object motion, and depth-of-field effects from a single reference image.
By decomposing camera-related signals into camera, world, and object coordinate spaces and unifying them through cross-coordinate conditioning fusion, our approach preserves the physical meaning of each control signal while enabling coherent, disentangled, and cinematographically meaningful video generation.
Extensive experiments demonstrate consistent improvements over prior methods in camera accuracy, motion stability, and editability, without compromising visual quality.
Overall, this work highlights the importance of explicit multi-coordinate camera modeling for controllable and physically meaningful video synthesis, and provides a flexible foundation for camera-aware generative tools in cinematic content creation and film production.

\begin{figure*}[p]
\centering
\includegraphics[width=0.84\textwidth]{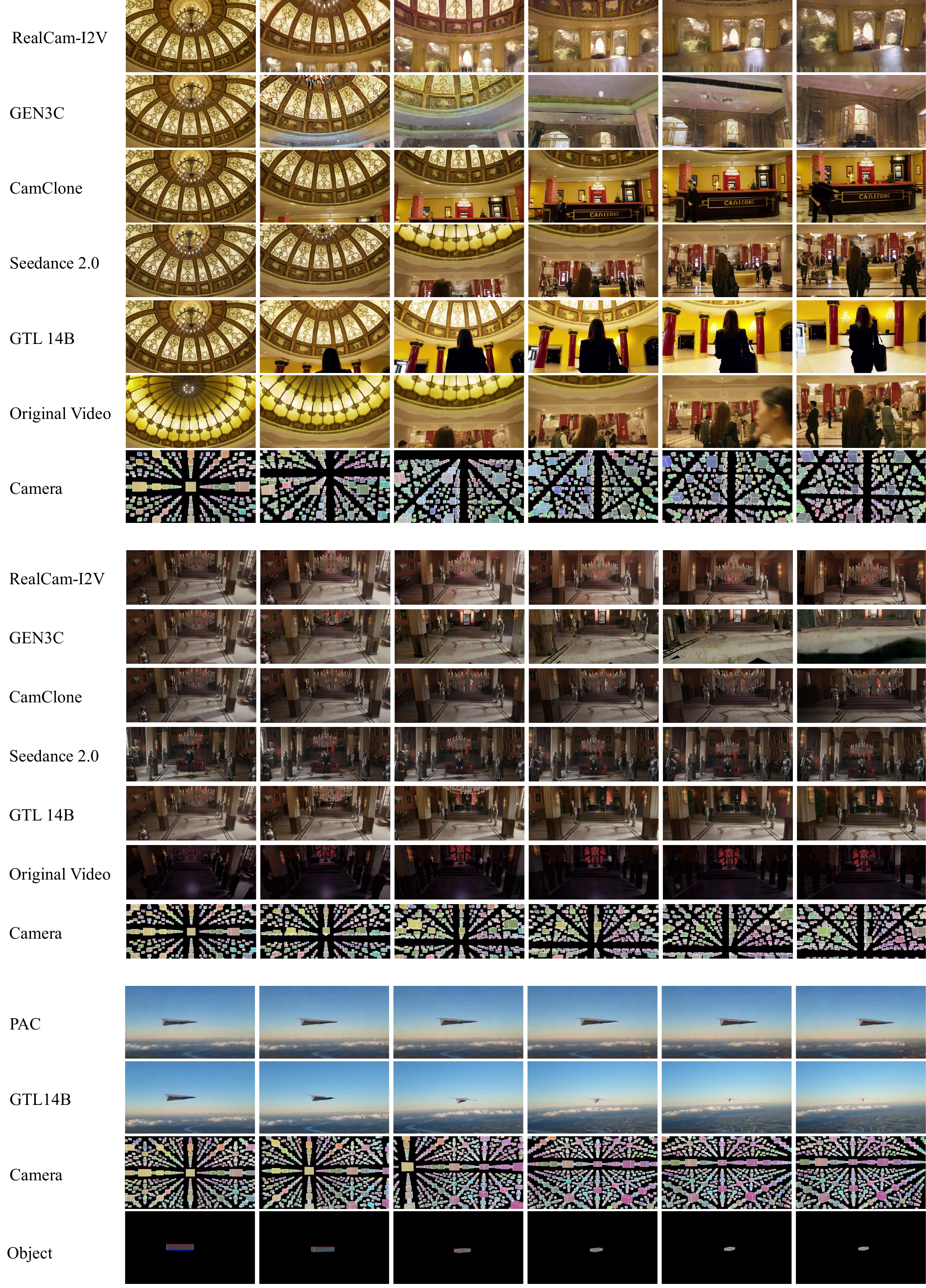}
\caption{
 Additional qualitative results corresponding to the experiments.
 Text prompts are omitted in this figure and provided in the Appendix Sec.~\ref{sec:add_prompt}.
}
\label{fig:add_figure}
\end{figure*}

\clearpage
\bibliographystyle{ACM-Reference-Format}
\bibliography{reference}

\clearpage
\appendix
\section{Additional Details on Geometry-Aware Conditioning}
\label{sec:appendix_conditions}

This appendix provides implementation details for the label prompt, camera motion condition, object motion condition, depth-of-field condition, and zoom augmentation.

\subsection{Label Prompt Details}
\label{app:prompt}

To ensure disentaglement between camera control and content description, we explicitly constrain the captions to exclude any camera related information.
We use the following prompt for dense video caption generation with InternVL2.5:

\noindent\fbox{
\begin{minipage}{0.95\linewidth}
\ttfamily
{\scriptsize
You are a professional video description master, your task is to carefully observe the input video, and give a detailed and dense description of the video, please follow the instructions below: 1. It is required to pay attention to the timing of the video, reflecting the development process of the events in the video over time and the details. 2. The output content should not include camera-related information, such as camera shooting angles and camera movements. 3. Require the output to capture as much video content as possible, including details, considering both completeness and specificity. It should provide comprehensive information that is tied to specific elements of the video. 4. The output description is required to be accurate and coherent, without misinformation and misunderstanding. The output must be factually accurate, coherent in content description, and consistent with the video content. 5. Do not give the info about camera such as camera shooting Angle and camera Movement. The Output should no more than 225 words.
}
\end{minipage}
}

\subsection{Camera Modeling and Camera Motion Condition}
\label{sec:appendix_camera}

\paragraph{Metric-Scale Camera Recovery.}
We combine Pi3~\cite{wang2025pi} and Moge2~\cite{wang2025moge} to recover metric-scale camera motion.
Pi3 provides temporally stable camera poses and scene structure up to an unknown global scale, while Moge2 predicts per-frame metric depth with limited temporal consistency.

To estimate the global scale, we randomly sample point pairs $(\mathbf{x}_a, \mathbf{x}_b)$ from reliable image regions, back-project them into 3D using Pi3 and Moge2, and compute distance ratios
$d^{\text{Moge}}_k / d^{\text{Pi3}}_k$.
We discard the top and bottom $20\%$ of ratios both within each frame and globally across the video, and average the remaining values to obtain a single video-level scale factor $s$.

After scaling Pi3 translations and associated 3D structure, we apply a Gaussian temporal filter with a window size of $0.5$ seconds independently to translation and rotation sequences.

For camera intrinsics, we use MapAnything~\cite{keetha2025mapanything} as an initial estimator.
Its intrinsic scale is aligned to the recovered metric camera space using the same pairwise-distance scaling strategy, and further adjusted to account for rendering resolution.

\paragraph{Camera Motion Condition.}
We discretize the surrounding 3D space using a regular grid of axis-aligned cubes.
Cubes are placed at uniform intervals of $0.8$ meters along the $x$, $y$, and $z$ axes in world coordinates, and each cube has an edge length of $0.15$ meters.
Only cubes within a distance range of $[1.5, 10]$ meters from the camera center are considered visible.
When more than 300 cubes are visible, we render only the 300 closest cubes.

All camera poses are normalized to the first frame.
The visible cubes are rendered using the recovered camera intrinsics and extrinsics.

To encode spatial position, each cube is assigned a color from a periodic spatial color field with a period of $10$ meters along each axis.
Specifically, we randomly sample $N=269$ anchor points
$\{(\mathbf{p}_i, \mathbf{c}_i)\}_{i=1}^N$ in a unit cube $[0,1]^3$,
where $\mathbf{p}_i \in \mathbb{R}^3$ satisfy
$\|\mathbf{p}_i - \mathbf{p}_j\|_2 > 0.12$ for all $i \neq j$
under periodic boundary conditions,
and $\mathbf{c}_i \in [0,255]^3$ are RGB colors.
The anchor set is tiled periodically in $[-1,2]$ along each axis.

For a 3D query point $\mathbf{x}$, we find its $k=10$ nearest anchors with distances $\{d_i\}_{i=1}^k$ and compute the interpolated color as
\begin{equation}
w_i = \frac{1}{d_i+\varepsilon} - \frac{1}{d_{\max}+\varepsilon}, \qquad
\tilde{w}_i = \frac{w_i}{\sum_{j=1}^k w_j},
\end{equation}
\begin{equation}
\mathbf{c}(\mathbf{x}) = \sum_{i=1}^k \tilde{w}_i \, \mathbf{c}_i .
\end{equation}

We approximate this dense color field using a multi-grid hash encoding~\cite{mueller2022instant} with a lightweight MLP, which is then used to paint all cubes efficiently.

\subsection{3D Bounding Box Modeling and Object Motion Condition}
\label{sec:appendix_object}

We use TokenHMR~\cite{dwivedi2024tokenhmr} to recover identity-consistent human meshes across frames.
For each tracked individual, we extract torso vertices from the recovered SMPL mesh~\cite{loper2023smpl} and fit a tight 3D bounding box.

The object motion condition is rendered as a projected 3D bounding box on a black background.
Pixels inside the projected box are assigned grayscale values in the range $[25,255]$ using per-video relative depth normalization, while background pixels remain at $0$.

During training, if no object motion is present, we use an all-black video as the object motion condition.

During inference, we use the same per-video depth normalization rule as in training.
The near bound is set to the minimum of $1$ meter and the closest distance from the camera to any point on the object bounding box.
The far bound is set to the maximum of $10$ meters and the farthest distance from the camera to the bounding box.
Depth values inside the box are normalized to this range and mapped to grayscale intensities.

To encode orientation, the 12 edges of the 3D bounding box are rendered using distinct colors sampled uniformly from the hue channel of HSV space.
The color assignment is fixed across frames and the dataset.

\subsection{Depth-of-Field Modeling and Depth-of-Field Condition}
\label{sec:appendix_dof}

We use the same physically motivated bokeh operator for both bokeh video construction and depth-of-field condition generation.
Given a per-frame focus depth, the operator computes blur radius from the depth deviation to the focal plane.
To stabilize blur near large depth discontinuities, we identify regions with abrupt depth changes and apply spatially adaptive smoothing to the blur-radius field.

The depth-of-field condition adopts a three-channel layered representation based on a synthetic checkerboard image.
All channels share the same initial black-and-white checkerboard pattern, where the image width is divided into 30 vertical segments and the height is adjusted to maintain approximately square cells.

Each channel corresponds to a canonical background depth in world coordinates, fixed at 5 meters, 15 meters, and 25 meters.
For each frame, foreground pixels inside the projected 3D bounding box are assigned the object’s metric depth, while background pixels retain the channel-specific depth.

The bokeh operator is applied independently to each depth map, producing three grayscale videos that are stacked into RGB channels to form the final depth-of-field condition video.
Because blur depends jointly on depth and bokeh strength, the resulting condition exhibits color variation across channels.

During training, if no bokeh effect is present, we use the original unblurred checkerboard as the condition.
During inference, the same procedure can be applied with manually specified bounding boxes to enable user-controlled depth-of-field effects.

\subsection{Zoom Augmentation}

We simulate zoom using temporal center cropping followed by resizing to the original resolution.
Crop factors and zoom intervals are sampled randomly.
The same zoom transformation is applied consistently to the camera motion condition video, object motion condition video, and depth-of-field condition video to preserve pixel-wise alignment.

\section{Evaluation Metrics}
\label{sec:metrics_appendix}

This appendix provides detailed definitions of all evaluation metrics used in the main paper.

\subsection{Camera Translation Error}
To quantify camera path accuracy in a scale-invariant manner, we compute the average relative translation error between predicted and ground-truth camera positions.
All camera trajectories are first normalized with respect to the camera pose at frame $t=0$, such that translations $\mathbf{t}$ represent relative displacements from the initial camera position, and rotations $\mathbf{R}$ are defined relative to the initial orientation.

The translation error is defined as:
\begin{equation}
\mathcal{E}_{\text{trans}} =
\frac{1}{T}\sum_{t=1}^{T}
\frac{
\left\|
\mathbf{t}^{\text{pred}}_t - \mathbf{t}^{\text{gt}}_t
\right\|_2
}{
\max{(\max_{1 \leq \tau \leq T}
\left\|
\mathbf{t}^{\text{gt}}_\tau
\right\|_2,\ 1)}
} * 100,
\end{equation}
The error metric is summarized as a single scalar value.

\

\subsection{Camera Rotation Error}
Camera orientation accuracy is evaluated using the geodesic distance between predicted and ground-truth rotations:
\begin{equation}
\mathcal{E}_{\text{rot}} =
\frac{1}{T}\sum_{t=1}^{T}
\arccos\left(
\frac{\operatorname{trace}\left(\mathbf{R}^{\text{pred}}_t
\mathbf{R}^{\text{gt}\top}_t\right)-1}{2}
\right),
\end{equation}
where $\mathbf{R}^{\text{pred}}_t$ and $\mathbf{R}^{\text{gt}}_t$ denote rotation matrices at frame $t$.
The error is reported in degrees.

\subsection{Mask Intersection-over-Union (mIoU)}
Foreground spatial consistency is evaluated using mean Intersection-over-Union (mIoU):
\begin{equation}
\text{mIoU} =
\frac{1}{T}\sum_{t=1}^{T}
\frac{|M^{\text{pred}}_t \cap M^{\text{gt}}_t|}
{|M^{\text{pred}}_t \cup M^{\text{gt}}_t|},
\end{equation}
where $M^{\text{pred}}_t$ and $M^{\text{gt}}_t$ denote predicted and reference foreground masks at frame $t$, obtained using SAM2.

\subsection{3D Bounding Box Pose Error}
We evaluate object-level geometric stability using both translation and rotation errors of the 3D bounding box. We employ a greedy algorithm to match $M$ trajectory pairs ($M=2$ in our setting), computing the error exclusively on frames $t\in \mathcal{T}_{\text{comm}}$ where $M$ bounding boxes are all detected.

The screen-space center deviation is defined as:
\begin{equation}
\mathcal{E}_{\text{bbox-trans}} =
\frac{1}{M\cdot\left|\mathcal{T}_{\text{comm}}\right|}\sum_{m=1}^{M}\sum_{t\in\mathcal{T}_{\text{comm}}}
\frac{
\left\|
\mathbf{c}^{\text{pred}}_{t,m} - \mathbf{c}^{\text{gt}}_{t,m}
\right\|_2
}{
\sqrt{H^2 + W^2}
},
\end{equation}
where $\mathbf{c}_t$ is the projected 2D center of the bounding box, and $H$, $W$ are image height and width.

Similarly, bounding box rotation error is measured using geodesic distance:
\begin{equation}
\begin{split}
\mathcal{E}_{\text{bbox-rot}} = \, &
\frac{1}{M\cdot\left|\mathcal{T}_{\text{comm}}\right|} \\
& \sum_{m=1}^M\sum_{t\in\mathcal{T}_{\text{comm}}}
\arccos\left(
\frac{\operatorname{trace}\left(\mathbf{R}^{\text{pred}}_{\text{bbox},t,m}
\mathbf{R}^{\text{gt}\top}_{\text{bbox},t,m}\right)-1}{2}
\right).
\end{split}
\end{equation}

\subsection{Trajectory Consistency Error}
Long-term motion fidelity is evaluated using trajectory consistency error computed from point tracks extracted by CoTracker:
\begin{equation}
\mathcal{E}_{\text{traj}} =
\frac{1}{N T}\sum_{i=1}^{N}\sum_{t=1}^{T}
\frac{
\left\|
\mathbf{p}^{\text{pred}}_{i,t} - \mathbf{p}^{\text{gt}}_{i,t}
\right\|_2
}{
\sqrt{H^2 + W^2}
},
\end{equation}
where $\mathbf{p}_{i,t}$ denotes the 2D position of point $i$ at frame $t$.

\subsection{Visual Quality and Semantic Alignment}
We evaluate perceptual quality using Fr\'echet Inception Distance (FID) and Fr\'echet Video Distance (FVD).
Semantic alignment between generated videos and input prompts is evaluated using the Qwen VQA Score, computed by querying a vision--language model on sampled frames and averaging the confidence scores.

\section{Inference hyperparameters}
\label{sec:hyperparameters}
During inference, we used the unipc \cite{zhao2023unipcunifiedpredictorcorrectorframework} sampling method, with the number of steps set to 20. The classifier-free guidance scale is set to 6.

\section{Additional Ablation Study}
\label{sec:add_ablation}

To evaluate the effectiveness of our camera extraction pipeline, we replace our metric-scale camera parameters with those directly estimated by MapAnything. 
We then train a GTL-1.3B model using these parameters as-is, while keeping all other components and training settings identical to the main model.

We observe that MapAnything produces camera parameters with inconsistent intrinsic across frames, leading to noticeable instability in the resulting camera motion condition.
This inconsistency introduces additional difficulty during training and degrades controllability.

As shown in Table~\ref{tab:ablation_dataset}, the model trained with our camera pipeline achieves significantly better performance in camera-related metrics compared to the MapAnything baseline, demonstrating the superiority of our camera extraction pipeline for controllable camera modeling.

\begin{table}[t]
\centering
\setlength{\tabcolsep}{6pt}
{\scriptsize
\begin{tabular}{lccc}
\toprule
Camera Extraction Method
& \begin{tabular}[c]{@{}c@{}}Cam Trans.\\ Err.  $\downarrow$\end{tabular}
& \begin{tabular}[c]{@{}c@{}}Cam Rot.\\ Err. (°) $\downarrow$\end{tabular} \\
\midrule
MapAnything  &  13.53   &  6.84   \\
\textbf{GTL}  & \textbf{10.15}  & \textbf{5.07} \\
\bottomrule
\end{tabular}
}
\caption{Ablation study on camera extraction pipeline.
We compare camera parameters directly estimated by MapAnything with those produced by our pipeline under the same training setting.
The results show that our camera extraction pipeline provides more reliable supervision, leading to improved camera translation and rotation accuracy.
}
\label{tab:ablation_dataset}
\end{table}

\section{Text Prompts for Experiments}

We provide the text prompts corresponding to all qualitative results for reproducibility.
For fairness, all methods share the same prompts, except for camera-related descriptions, which are ignored by our model but used by baseline methods when applicable.
Camera-related descriptions are marked with [Camera].

For Seedance 2.0, we use a different prompt format that incorporates both the input image and reference camera motion.
Specifically, the prompt follows the template:
“Use the input image as the first frame, follow only the camera motion from the reference video, and generate a new video described as: <PROMPT>.”,
where <PROMPT> corresponds to the descriptions listed below.

\paragraph{Camera Experiments (Fig.~\ref{fig:main_figure}).}
\label{sec:camera_prompt}
\begin{itemize}
    \item During the day, sunlight streamed in through the windows. Indoors, a man in a white suit, yellow tie, and glasses, with his hair meticulously styled, held a large bouquet of red roses with a serious expression.  [\textbf{Camera}: The camera began with an overhead shot from above the ceiling, then rotated downwards, finally focusing on the pink ribbon decorating the rose bouquet.]
\end{itemize}

\paragraph{Bounding Box Experiments (Fig.~\ref{fig:bbox_figure}).}
\label{sec:bbox_prompt}
\begin{itemize}
    \item During the daytime, in a European-style residential complex, a red sedan slowly drives away from the complex's gate, traveling along a gray road. The complex is surrounded by two-story European-style residential buildings. A row of bicycles is parked to the left of the building entrance, and a motorcycle is parked to the right.  [\textbf{Camera}: The camera follows the car as it moves to the right, and when the car reaches the complex gate, the camera stops moving and begins to pan to the right until the car drives out of the complex gate.]
    \item The video begins with a medium shot of a bright, modern office lobby. In the foreground is a professional woman with her hair tied back. Standing slightly behind her is a man wearing a black beanie. Initially, the woman has a warm and professional smile, while the man's expression is more reserved. The woman walks forward, always facing the camera, with the man following closely behind. The woman guides the man through the office space. As they move forward, the camera smoothly moves backward, keeping them centered in the frame. More environmental details gradually become visible, including a marble elevator and staircase in the background. 
\end{itemize}

\paragraph{Disentanglement  Experiments (Fig.~\ref{fig:disentangle}).}
\label{sec:disentangle_prompt}
\begin{itemize}
    \item A man on a city street, surrounded by buildings.
\end{itemize}

\paragraph{Depth-of-Field Experiments (Fig.~\ref{fig:bokeh_figure}).}
\label{sec:bokeh_prompt}
\begin{itemize}
    \item This video clip documents an encounter between two men on a European street. The scene opens with a medium shot. In the foreground, on the right, a man is looking at the camera. In the background, another man is walking away down the street. The man in the foreground turns around and begins to walk forward. As he walks forward, the man he was following stops, turns around to face him, and stands there expressionlessly.
    \item A woman is talking to a girl.
\end{itemize}

\paragraph{Hitchcock Experiments (Fig.~\ref{fig:hitchcock_figure}).}
\label{sec:hitchcock_prompt}
\begin{itemize}
    \item A man standing on a green grassy field, with majestic mountains in the background under a dim, overcast sky.
\end{itemize}

\paragraph{Additional Experiments (Fig.~\ref{fig:add_figure}).}
\label{sec:add_prompt}
\begin{itemize}
    \item A woman with long brown hair, wearing a black suit and carrying a large black shoulder bag, walks slowly towards the hotel reception desk with her back to the camera. The lobby is predominantly yellow, with crystal chandeliers hanging from the ceiling, and red load-bearing columns with gold accents symmetrically positioned on either side of the reception desk. [\textbf{Camera}: Starting from the rooftop, the camera moves downward into the interior, then smoothly pushes forward through the hall]
    \item Two rows of armored warriors stand motionless on either side. In the center of the main reception area sits a man in a suit, exuding an imposing presence. [\textbf{Camera}: The camera starts slightly facing the ground, then levels to a forward view and continues a steady forward push]
    \item A paper airplane floated in the air, slowly drifting away into the distance. [\textbf{Camera}: The camera remains stationary and rotates to the right]
\end{itemize}

\end{document}